\pdfoutput=1

\documentclass[11pt]{article}

\usepackage[]{emnlp2021}
\usepackage{bm}
\usepackage{times}
\usepackage{latexsym}
\usepackage{amsmath}
\usepackage{algpseudocode}
\usepackage{graphics}
\usepackage{epsfig}
\usepackage{array}
\usepackage{amssymb}
\usepackage[T1]{fontenc}

\usepackage[utf8]{inputenc}

\usepackage{microtype}

\title{Which is Making the Contribution: Modulating Unimodal and Cross-modal Dynamics for Multimodal Sentiment Analysis}

 \author{Ying Zeng$^1$ ,
  Sijie Mai$^1$,
  Haifeng Hu\\
         Sun Yat-sen University \\  \{zengy268,maisj\}@mail2.sysu.edu.cn, \ huhaif@mail.sysu.edu.cn}

\begin{document}
\maketitle

\begin{abstract}

Multimodal sentiment analysis (MSA) draws increasing attention with the availability of multimodal data. The boost in performance of MSA models is mainly hindered by two problems. On the one hand, recent MSA works mostly focus on learning cross-modal dynamics, but neglect to explore an optimal solution for unimodal networks, which determines the lower limit of MSA models. On the other hand, noisy information hidden in each modality interferes the learning of correct cross-modal dynamics. To address the above-mentioned problems, we propose a novel MSA framework \textbf{M}odulation \textbf{M}odel for \textbf{M}ultimodal \textbf{S}entiment \textbf{A}nalysis ({$ M^3SA $}) to identify the contribution of modalities and reduce the impact of noisy information, so as to better learn unimodal and cross-modal dynamics. Specifically, modulation loss is designed to modulate the loss contribution based on the confidence of individual modalities in each utterance, so as to explore an optimal update solution for each unimodal network. Besides, contrary to most existing works which fail to explicitly filter out noisy information, we devise a modality filter module to identify and filter out modality noise for the learning of correct cross-modal embedding. Extensive experiments on publicly datasets demonstrate that our approach achieves state-of-the-art performance.

\end{abstract}

\section{introduction}
The availability of multimodal data enables us to perform many downstream tasks with cross-modal information, such as conversation generation, multimodal sentiment analysis, etc. In the field of sentiment analysis (MSA), recently researchers leverage the rich information contained in different modalities (e.g., audio, visual, language) to design multimodal models, and existing works mainly focus on exploring cross-modal dynamics and designing sophisticated fusion methods \cite{ARGF,MCTN,Poria2017A,MISA,mai2021unimodal}. 

While existing MSA models are mostly optimized by multimodal loss, the design towards the optimization of unimodal networks in MSA models is often neglected. However, the reach of optimal unimodal networks determines the lower limit of the whole MSA models, which should specifically addressed for the higher performance of the models. Besides, an optimal solution for each modality also ensures the performance of MSA models even with the absence of any modality.

Moreover, even with satisfactory unimodal networks, it is not always the case that multimodal models reach higher performance than the unimodal ones \cite{TASLP}. The reason may be that, a modality may not contain useful information in some utterances and may even carry noises, which hinders the learning of correct multimodal embedding. Some attention-based methods leverage attention mechanism to determine modality importance \cite{CIA, MTL}, which can filter out noise information in a certain degree, but those methods introduce a large amount of parameters and increase the risk of overfitting. Moreover, despite the attention on informative modalities, the noisy modalities cannot be explicitly filtered out. 


Based on the aforementioned problems, we mainly concern about two questions: how to obtain an optimal unimodal network; which modality is informative and how to filter out noisy modalities. We hold the intuition that each modality carries modality-specific information, whose importance varies from one another. Moreover, the role of the same modality also varies (the amount of useful and noisy information varies in different utterances).
To address these concerns, we propose a novel \textbf{M}odulation \textbf{M}odel for \textbf{M}ultimodal \textbf{S}entiment \textbf{A}nalysis {$ M^3SA $} to modulate the training of different modalities.

Specifically, modulation loss and modality filter module are designed to identify import modalities and reduce the negative impact of noisy information. To learn an optimal unimodal network, modulation loss is proposed to modulate the training of each unimodal network. The core idea is that during the training stage, the modulation function manages to modulate the loss contribution of each modality according to the confidence of all the modalities \cite{focal}, which enables the model to balance multi-modal information and identify the importance of each modality at each utterance. In this way, the model can dynamically adjust the contribution from different modalities so as to better leverage the importance information hidden within each modality to update the unimodal networks. With our proposed modulation loss, the training of individual unimodal networks is modulated and they can be better optimized by reducing the inference of the noisy modalities at each utterance.

Besides, to obtain correct multimodal embedding, we design a modality filter module (MFM) to identify modality importance and explicitly filter out noisy modalities. We present two possible candidates of the filter of MFM, i.e., a hard-filter and a soft-filter, where the hard-filter provides a binary choice $\{0, 1\}$ to retain or filter out individual modalities, while the soft-filter outputs a number between [0, 1] to filter out noisy information based on the noise level. Moreover, instead of directly removing the noisy modalities or tokens \cite{Chen2018Multimodal,rl2}, inspired by \cite{mask}, we innovative to train a baseline embedding for each modality and replace the noisy embedding with it, such that our method can be fitted into any fusion mechanisms and compensate for the loss of unimodal information.

In brief, the contributions can be summarized as:

\begin{itemize}
\item We propose a novel framework $ M^3SA $ to modulate the training of MSA models, which aims to explore optimal solution for unimodal networks and multimodal embedding.
\item A cross-modal modulation loss is devised to modulate the contribution of each modality based on the confidence of individual modalities during the training stage, and it can reduce the interference from noisy modalities so that unimodal networks can be better optimized, which is often neglected in existing works.
\item A modality filter module (MFM) is designed to identify noisy modalities and filter them out where soft-filter, hard-filter and unimodal embedding baselines are proposed, so as to minimize the negative impact of noisy information and obtain correct multimodal embedding. Compared with attention-based methods, MFM introduces much less parameters and can explicitly filter out noisy modalities.
\item Our proposed method is compared with several models on public datasets and achieves state-of-the-art performance, which demonstrates its effectiveness and superiority.
\end{itemize}

\section{Related Work}

In the field of MSA, each sample is an utterance that captures different views with complementary information. Most previous works focus on elaborately designing various fusion strategies so that the model can explore inter-modal dynamics to sufficiently learn a joint embedding, including simple ways like early fusion and late fusion \cite{Wollmer2013YouTube, Rozgic2012Ensemble, Poria2017Convolutional, Poria2017Context}, and more advanced fusion strategies like tensor-based fusion \cite{Liu2018Efficient, Zadeh2017Tensor, HFFN}, graph fusion \cite{ARGF,MOSEI,mai2020analyzing}, factorization methods \cite{MFM,MMB}, fine-tuning BERT \cite{MAG-BERT, CM-BERT} etc.


The above-mentioned methods focus on exploring more advanced fusion strategies, and optimize the whole network mostly based on multimodal loss so as to achieve higher performance for MSA task. While more attention is paid on the optimization of multimodal networks, specifically designed method for optimizing individual unimodal networks is neglected. 
We hold that apart from the learning of cross-modal dynamics, it is also important to reach an optimal solution for the optimization of unimodal networks. To achieve this goal, we specifically design a modulation loss to modulate the loss contribution of unimodal networks based on their confidence. We train all unimodal networks with the modulation loss across all data points with the aim to reaching optimal parameters on the corresponding dataset.

\begin{figure*}[h]
  \centering
  \includegraphics[width=0.95\linewidth]{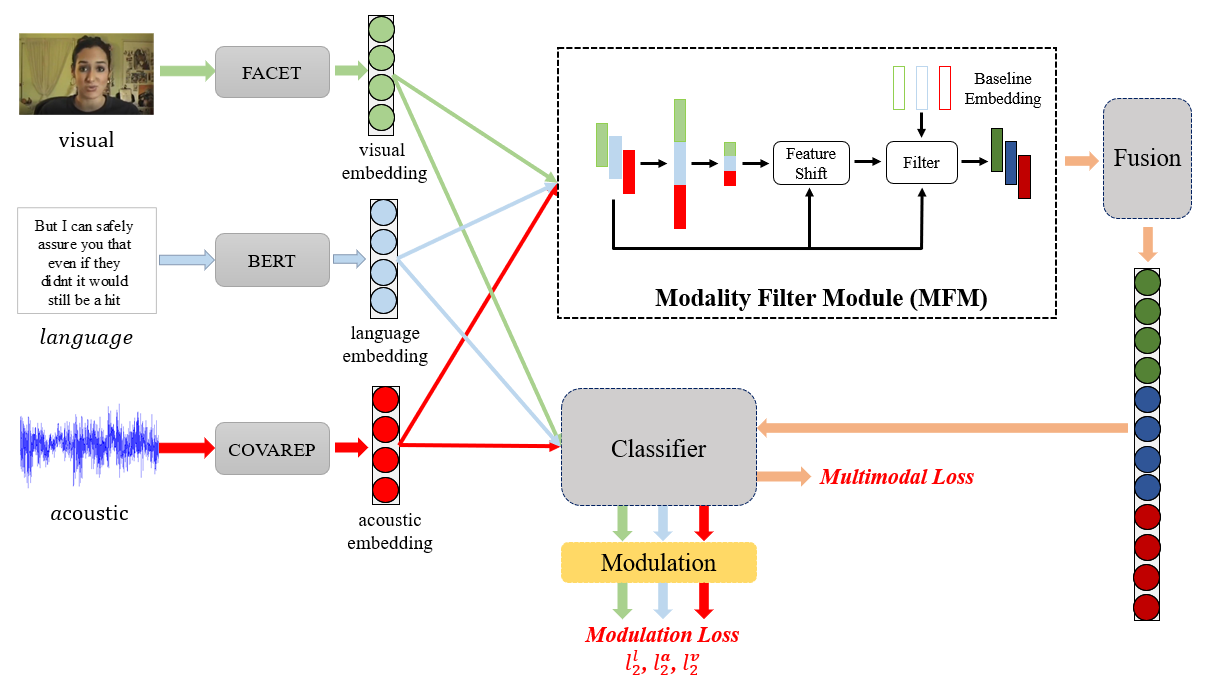}
  \caption{\label{diagram}The diagram of our proposed $M^3SA$.
  }
\end{figure*}



Another problem in the field of MSA is the interference between modalities. 
Noisy modalities can interfere the learning of other modalities and the correct multimodal embedding. Some \textbf{attention-based fusion} methods such as Context-aware Interactive Attention (CIA) \cite{CIA}, Multi-Task Learning (MTL) \cite{MTL} and Multilogue-Net \cite{Multilogue-Net} that apply cross-modal attention mechanism consider the importance of different modalities and assign different weights to them. But they focus on identifying and highlighting important modalities, and can not explicitly filter out noisy modalities. Although these models have considered modality importance, we format it from a different perspective instead of learning attention weights. Specifically, we focus on identifying and filtering out noisy modalities with a modality filter module (MFM), which introduces much few parameters than attention mechanisms and can explicitly filters out noisy information. Actually, there also exists works that aim to filter out the noisy modalities or the tokens within modality using reinforcement learning (RL) \cite{Chen2018Multimodal,rl2}. However, RL is unstable in training and suffers from high variants and control variates that requires auxiliary models or multiple evaluations of the network \cite{hard1,hard3}. Moreover, they provide a binary choice to retain or filter out the whole noisy modality, and modality-specific information may be lost. Unlike it, our proposed MFM is much more easier to train, and at the same time MFM considers the baseline embedding to compensate the loss of modality-specific unimodal information.



\section{Algorithm}

\subsection{Notations and Problem Formulation}
Our task is to perform multimodal sentiment analysis with multimodal data by scoring the sentiment intensity. The input to the model is an utterance \cite{Olson1977From} (i.e., a segment of a video bounded by pauses and breaths), each of which has three modalities, i.e., acoustic ($a$), visual ($v$), and language ($l$). The sequences of acoustic, visual, and language modalities are denoted as $\bm{u^a} \in \mathbb{R}^{T_a \times d_a}$, $\bm{u^v} \in \mathbb{R}^{T_v \times d_v}$, and $\bm{u^l} \in \mathbb{R}^{T_l \times d_l}$, where $T_a$, $T_v$ and $T_l$ represent the length of the audio, visual and language modality, respectively, and $d_a$, $d_v$ and $d_l$ denote the dimensionality of the audio, visual and language modality, respectively.

\subsection{Overall Algorithm}

Formally, a traditional multimodal learning system can be formulated as:
\begin{equation}
\setlength{\abovedisplayskip}{0pt}
\setlength{\belowdisplayskip}{0pt}
  \bm{x}^{m}=\bm{F}^m(\bm{u}^m; \theta_m), m \in \{ l, a, v\}
\end{equation}
\begin{equation}
\setlength{\abovedisplayskip}{0pt}
\setlength{\belowdisplayskip}{0pt}
  y_M = \bm{F}^M(\bm{x}^{l}, \bm{x}^{a}, \bm{x}^{v}; \theta_M)
\end{equation}
where $y_M$ is the prediction, $\bm{F}^m$ parameterized by $\theta_m$ and $\bm{F}^M$  parameterized by $\theta_M$ refer to the unimodal and multimodal network, respectively. $\bm{U}^m\in \mathbb{R}^{T_m \times d_m}$ is the input raw feature of modality $m$ where $T_m$ is the sequence length.
To update the parameters of the multimodal system, we have the following equation:
\begin{equation}
\setlength{\abovedisplayskip}{3pt}
\setlength{\belowdisplayskip}{3pt}
 \ell = ||y-y_M||_1,\ \theta\!\leftarrow\! \theta - \alpha \frac{\partial\ell}{\partial \theta}
\end{equation}
where $y$ is the ground truth label, $\theta \in \{\theta_a,\theta_v,\theta_l,\theta_M \}$, $\alpha$ is the learning rate, and $\ell$ is mean absolute error (MAE).


Unlike the traditional multimodal learning system which mostly focuses on optimizing the whole multimodal framework, we decouple the learning procedure of unimodal and multimodal networks, introduce modulation losses to specifically optimize the unimodal networks for learning better unimodal representations, and design modality filter module (MFM) for identifying and filtering out noisy modalities. As illustrated in Fig.~\ref{diagram}, given an input utterance of three modalities, we first obtain the unimodal representations via unimodal networks. Modulation loss is specifically designed to train individual unimodal networks by modulating the loss contributions of each modality.
Besides, the output of each unimodal network will be sent to the MFM, and in this way, noisy modalities can be identified and filtered out. With our proposed method, we can modulate the learning of correct unimodal and multimodal dynamics, and minimize the negative impact of noisy information. In a word, our multimodal learning system is formulated as:
\begin{equation}
\setlength{\abovedisplayskip}{0pt}
\setlength{\belowdisplayskip}{0pt}
\label{eq4}
  \bm{x}^{m}=\bm{F}^m(\bm{u}^m; \theta_m), m \in \{ l, a, v\}
\end{equation}
\begin{equation}
\setlength{\abovedisplayskip}{0pt}
\setlength{\belowdisplayskip}{0pt}
\label{eq5}
  y_{m}=\bm{C}(\bm{x}^{m}; \theta_c), \ \ell^m = \left|y_{m}-y\right|
 \end{equation}
 \begin{equation}
\setlength{\abovedisplayskip}{0pt}
\setlength{\belowdisplayskip}{0pt}
\label{eq5555}
  l^{m}_2=\text{Modulation}(\ell^a, \ell^v, \ell^l; \ell^m)
 \end{equation}
  \begin{equation}
\setlength{\abovedisplayskip}{0pt}
\setlength{\belowdisplayskip}{0pt}
\label{eq555}
  \bm{x}^{m}_2=\text{MFM}(\bm{x}^{l}, \bm{x}^{a}, \bm{x}^{v}; \bm{x}^{m})
 \end{equation}
 \begin{equation}
\setlength{\abovedisplayskip}{0pt}
\setlength{\belowdisplayskip}{0pt}
  \bm{x^M} = \bm{F}^M(\bm{x}^{l}_2, \bm{x}^{a}_2, \bm{x}^{v}_2; \theta_M)
\end{equation}
\begin{equation}
\setlength{\abovedisplayskip}{0pt}
\setlength{\belowdisplayskip}{0pt}
\label{eq6}
  y_M=\bm{C}(\bm{x^M}; \theta_c), \ \ell_M = \left|y-y_M\right|
 \end{equation}
where $\bm{C}$ is the classifier that  takes encoded representation as input and outputs the sentiment prediction, which  is shared across unimodal and multimodal networks to force the learned unimodal and multimodal representations to have approximately same distributions. As illustrated in Eq.~\ref{eq5555}, the unimodal losses are adjusted by a Modulation function, which helps to identify the contribution of each modality of the current utterance to the optimization of the respective unimodal network. $ l^{m}_2$ is used to update the respective unimodal network. Moreover, in Eq.~\ref{eq555}, MFM is introduced to identify and replace the uninformative modalities with the learned unimodal baseline embeddings to filter out the noisy information that interferes the learning of the cross-modal interactions. The detailed introduction of the modulation function and the MFM is shown in Section 3.3 and 3.4, respectively. 

Unlike most existing works which need sophisticated designed fusion methods to sufficiently explore cross-modal dynamics, our proposed $M^3SA$ can leverage simple fusion method to reach the state-of-the-art performance with  better generalization ability. Also note that our algorithm is model-agnostic, and we can integrate any sequence learning networks into our unimodal networks $\bm{F}^m$.  In this paper, we apply Transformer-based \cite{transformer} architectures to build up the unimodal networks. As for the multimodal network $\bm{F}^M$, we introduce different fusion mechanisms to evaluate the algorithm. Please refer to Appendix for the details about the unimodal and multimodal networks.


\subsection{Modulation Loss}
The cross-modal modulation function is proposed to modulate the loss contribution of
each modality as a function of the confidence of individual
modalities. This is based on the assumption that each modality carries various modality-specific information, whose importance varies from one modality to another modality. And in different utterances, the role of the same modality also varies (in some utterances, this modality is important, while in other utterance, it contains only the noisy information). Instead of learning the fixed attention weight for each modality as the previous methods do \cite{RAVEN,ARGF}, we seek to dynamically adjust the contribution from different modalities so as to better leverage the important information hidden within each modality to update the network, and effectively reduces the interference of the noisy utterances. Compared to the attention mechanism, the modulation loss directly has influence on the optimization procedure, which is more straightforward and non-parametric.

How do we dynamically determine the contribution of each modality during training? A intuitive idea is that we can estimate the value of the unimodal loss, under the assumption that the smaller the value of the unimodal loss, the more discriminative it is for the task, and a higher weight shall be assigned so as to better leverage the discriminative information hidden in this modality to update the network. More importantly, when assigning weight to each unimodal loss, we should have a global view on all the modalities to consider the value of the other unimodal losses to estimate the relative importance and adjust the weight for this modality accordingly. The modulation loss can be formulated as (taking language modality as an example):
\begin{equation}
\setlength{\abovedisplayskip}{0pt}
\setlength{\belowdisplayskip}{0pt}
\label{eq5}
  \ell^l_2 =\text{Modulation}(\ell^l, \ell^a, \ell^v; \ell^l)
 \end{equation}
where $\ell^l_2$ is the modulation loss for language modality.
The Modulation function aims to learn the weight for unimodal loss by estimating the discriminative information in all the modalities (this is why we call it modulation). The formulation of the Modulation function could have many choices. In practice, we formulate
it as:
\begin{equation}
\setlength{\abovedisplayskip}{0pt}
\setlength{\belowdisplayskip}{0pt}
\label{eqmean}
\alpha =\frac{1}{\frac{1}{3} \sum_{m}^{m\in\{l,a,v\}} \frac{1}{l^{m}}}=\frac{3}{\sum_{m}^{m\in\{l,a,v\}} \frac{1}{l^{m}}}
 \end{equation}
\begin{equation}
\setlength{\abovedisplayskip}{0pt}
\setlength{\belowdisplayskip}{0pt}
\label{eqmodulation}
  \alpha_l = \alpha \times \ell^a \times \ell^v
 \end{equation}
 \begin{equation}
\setlength{\abovedisplayskip}{0pt}
\setlength{\belowdisplayskip}{0pt}
\label{eq5}
  \ell^l_2 = \ell^l \times \alpha_l
 \end{equation}
where $\alpha$ is the harmonic mean of the three unimodal losses which performs a kind of scale on the weight of unimodal losses,  and $\alpha_l$ is the weight for the language loss.
By using the loss values of other modalities to compute weights for the current modality, the weight of the current modality reduces when the other modalities obtain relatively low losses (i.e., other modalities have high confidence for prediction). In other words, the modality that has a relatively high loss obtains a low weight when updating the corresponding unimodal network, which dynamically reduces the influence of noisy modalities to the network. This simple operation is shown to be very effective (see experiment).

\subsection{Modality Filter Module}

The problem of noisy modalities negatively affects the learning of other informative modalities and hinders higher performance of existing MSA models. 
Many existing works try to identify modality importance with attention mechanisms \cite{ARGF,RMFN}, which can highlight useful tokens or modalities and filter certain noisy information out. However, those methods cannot completely filter out the noisy information and only tend to assign high weight to the informative modalities. 
\citet{Chen2018Multimodal} leverage reinforcement learning (RL) to learn a gate controller for each modality, which can shut off noisy modalities. But RL suffers from high variance and introduces more parameters and optimization objective \cite{hard1}, which is unstable in training.

Unlike previous methods, we propose a modality filter module (MFM) to selectively filter noisy modalities out, in which way the negative impact of noisy information can be minimized. Unlike \cite{Chen2018Multimodal} which only considers non-lexical modalities as the possible noisy modalities, we aim to identify if the three modalities in each utterance contain noisy information, and if they should contribute to the final prediction.

Mathematically, the deployment of MFM firstly takes the feature embeddings of all the modalities as inputs, and calculates a feature shift of the overall multimodal embedding to each specific unimodal embedding, which can be formulated as:
  \begin{equation}
  \begin{aligned}
   & \bm{x^M} = \bm{x}^l\oplus\bm{x}^a\oplus\bm{x}^v \\
   & \bm{x'} = \text{Linear}(\bm{x^M}; \theta_L) \\
    \bm{x_{shift}^m} =& \text{ReLU}(\bm{x'-x^m}), m \in \{ l, a, v\}
    \end{aligned}
\end{equation}
where $\bm{x^M}$ denotes a multimodal representation by the concatenation of the embeddings of the three modalities, $\bm{x'}$ represents the processed multimodal representation which preserves the same dimensionality as individual modalities by a linear transformation, and $\bm{x_{shift}^m}$ is the feature shift of modality $m$ compared to $\bm{x'}$. By using all the unimodal embeddings to modulate and determine the noisy level of each specific modality, the model can have a global view on all the modalities and determine which is informative and which is not.

With the obtained feature shift of each modality, MFM filters out noisy information by a Filter:
  \begin{equation}
  \setlength{\abovedisplayskip}{3pt}
\setlength{\belowdisplayskip}{3pt}
    s^m= \text{Filter}(\bm{x_{shift}^m}; \theta_f), m \in \{ l, a, v\}
\end{equation}
where \text{Filter} parameterized by $\theta_f$ outputs $s^m$, which determines whether to filter the  modality $m$ out based on its noise level. The \text{Filter} is trained across all utterances, and it can identify and filter out noisy modality. The realization of \text{Filter} has many possibilities, and we put forward two candidates in Section~\ref{soft} and Section~\ref{hard}. After obtaining the output $s^m$ from the \text{Filter}, the final embedding of the modality $m$ can be determined:
  \begin{equation}
  \label{eqbb}
  \setlength{\abovedisplayskip}{3pt}
\setlength{\belowdisplayskip}{3pt}
    \bm{x_2^m}= s^m \cdot \bm{x^m} + (1-s^m) \cdot \bm{b^m}
\end{equation}
where $\bm{x_{out}^m}$ represents the final embedding of the  modality $m$, which  contains much less noisy information. $\bm{x_{out}^m}$ of individual modalities is then leveraged to learn a correct multimodal embedding for MSA task. Besides, we assume that filtering out too much information of the noisy modality may degrade the performance, for the model may lose modality-specific information. To compensate the modality-specific information of noisy modalities, we learn a baseline embedding $\bm{b^m}$ for each modality. The unimodal baseline embedding $\bm{b^m}$ is a critical part of our MFM, which is trained across multiple data points in the dataset. $\bm{b^m}$ is assumed to integrate the general distributions and properties of each modality, and therefore it can compensate the modality-specific information for fusion. Moreover, instead of directly removing the noisy modalities or tokens \cite{Chen2018Multimodal,rl2}, the unimodal baseline embedding enables our model to fit into any fusion mechanism such as tensor fusion or element-wise multiplication, providing more generalization ability.

With our proposed MFM, our model is capable of identifying and filtering out noisy modalities. In this way, our proposed model can dynamically retain informative modalities to modulate the learning of correct multimodal embedding for each utterance. Besides, to minimize the negative impact of the absence of modality-specific information, the learned baseline embedding $\bm{b^m}$ of each modality helps to sufficiently learn cross-modal dynamics.


\subsubsection{Soft Filter}
\label{soft}
To realize the Filter function, we first consider the soft filter mechanism whose output value is not binary. The procedure for soft filter is shown below:
\begin{equation}
  \setlength{\abovedisplayskip}{0pt}
\setlength{\belowdisplayskip}{0pt}
    \bm{z^m} = \text{FC}(\bm{x_{shift}^m}; \theta_{fc})
\end{equation}
\begin{equation}
  \setlength{\abovedisplayskip}{0pt}
\setlength{\belowdisplayskip}{0pt}
   s_i^m = \frac{e^{\lambda\cdot z_i^m}}{\sum_{j=1}^2e^{\lambda\cdot z_j^m}},\ \ \bm{s^m} = [s_1^m, s_2^m]
\end{equation}
\begin{equation}
  \setlength{\abovedisplayskip}{0pt}
\setlength{\belowdisplayskip}{0pt}
    l^p =  1 - (s_1^m -s_0^m)^2
\end{equation}
where $\lambda$ is the scale factor to widen the distance between the elements in $s$, $FC$ is the fully-connected network activate by ReLU, and $\bm{s^m}\in \mathbb{R}^{2}$ is the assignment vector that determines the noisy level of modality. $l^p$ is the penalty loss that encourages the elements of $\bm{s^m}$ to be close to 0 or 1. Nevertheless, the elements of $\bm{s^m}$ are not likely to be binary because they are continuous. But via the soft filter, the model can learn to estimate how much information in the modality can be filtered out instead of directly filtering out all the information, providing more fine-grained filtering effect. Since the output of soft-filter is a 2-dimensional vector, Eq.~\ref{eqbb} should be rewritten as:
  \begin{equation}
  \setlength{\abovedisplayskip}{3pt}
\setlength{\belowdisplayskip}{3pt}
    \bm{x_2^m}= s^m_1 \cdot \bm{x^m} + s^m_2 \cdot \bm{b^m}  ,\ \ s^m_1 + s^m_2 = 1
\end{equation}
Soft filter differs from attention mechanism in following aspects: 1)  introducing scale factor $\lambda$ and penalty loss $l^p$ to reach better filtering effect; 2) introducing the unimodal baseline embedding to compensate the filtered modality-specific information; 3) merely modifying the unimodal embedding and can be integrated with any fusion mechanisms.
\subsubsection{Hard Filter}
\label{hard}
The output of the hard filter, i.e., $s^m$, is a scalar that is either $0$ or $1$. However, due to the discrete nature of $s^m$, training this kind of framework using gradient-based optimization algorithm is intractable. To resolve this problem, we follow \cite{hard1} to use reparameterization trick \cite{hard2} to compute the unbiased and low variance gradients. Specifically, we utilize the Hard Concrete distribution introduced in \cite{hard1}, which is a mixed discrete-continuous distribution on the interval [0, 1]. Hard Concrete assigns a continuous probability to exact zeroes or ones, and meanwhile it allows continuous outcomes in the unit interval such that the gradient can be computed via the reparameterization trick. The computation of $s^m$ for hard filter is illustrated as follows:
\begin{equation}
\begin{aligned}
&z^m = \text{FC}(\bm{x_{shift}^m}; \theta_{fc})\\
 &  \hat{s}^m =\operatorname{Sigmoid}((\log \frac{u}{1-u} + z^m ) / \beta)\\
 &\bar{s}^m =\hat{s}^m \times (\zeta-\gamma)+\gamma  \\
 & s^m = 1\  \text{iff}\ \bar{s}^m>0.5\ \text{else}\ s^m = 0
\end{aligned}
\end{equation}
where $\beta$ is the temperature, $\zeta$ and $\gamma$ are the hyper-parameter to scale $s^m$, and $u \sim \mathcal{U}(0,1)$ ($\mathcal{U}$ denotes Gaussian distribution).
Compared to using RL \cite{Chen2018Multimodal,rl2} to obtain the exact binary weight, using the Hard Concrete distribution is much more simple and stable in training, with no additional optimization objectives or components introduced. Via the hard filter, the model can completely filter out the noisy modalities which cannot be realized by the attention mechanisms.  For more details about Hard Concrete distribution, please refer to \cite{hard1}.

\section{Experiment}

\begin{table}[t]
\centering
 \caption{ \label{t1}\textbf{ The comparison with baselines on CMU-MOSI.} Note that QMF and MISA do not provide the code so we present the result from their papers.}
\resizebox{.95\columnwidth}{!}{\begin{tabular}{c|c|c|c|c|c}
 \hline
    & Acc7 & Acc2 & F1 & MAE & Corr \\
 \hline
 EF-LSTM  & 31.6 & 75.8 & 75.6 & 1.053 & 0.613  \\
 LF-LSTM  & 31.6 & 76.4 & 75.4 & 1.037 & 0.620  \\
TFN \cite{Zadeh2017Tensor} & 32.2 & 76.4  & 76.3 & 1.017 & 0.604 \\
 LMF \cite{Liu2018Efficient} & 30.6 & 73.8  & 73.7 & 1.026 & 0.602 \\
 MFN \cite{Zadeh2018Memory} & 32.1 & 78.0  & 76.0 & 1.010 &0.635  \\
RAVEN \cite{RAVEN} & 33.8 & 78.8  & 76.9 & 0.968 & 0.667 \\
 MULT \cite{MULT} & 33.6 & 79.3  & 78.3 & 1.009 & 0.667 \\
 QMF \cite{Quantum} & 35.5 & 79.7  & 79.6 & 0.915 & 0.696 \\
 MAG-BERT \cite{MAG-BERT} & 42.9 & 83.5 & 83.5 & 0.790 & 0.769 \\
 \hline
 $M^3SA$ (Hard) & 45.5 & 85.3  & 85.3 & 0.730 & 0.793 \\
$M^3SA$ (Soft) & \textbf{46.4} & \textbf{85.7}  & \textbf{85.6} & \textbf{0.714} & \textbf{0.794} \\
 \hline
 \end{tabular}}
 \vspace{-0.3cm}
\end{table}%

\begin{table}[t]
\centering
 \caption{ \label{t2}\textbf{ The comparison with baselines on CMU-MOSEI.} Note that IMR cannot perform regression task so that MAE and Corr are not available. }

\resizebox{.95\columnwidth}{!}{\begin{tabular}{c|c|c|c|c|c}
 \hline
    & Acc7 & Acc2 & F1 & MAE & Corr \\
 \hline
 EF-LSTM  & 46.7 & 79.1 & 78.8 & 0.665 & 0.621\\
 LF-LSTM  & 49.1 & 79.4 & 80.0 & 0.625 & 0.655 \\
 TFN \cite{Zadeh2017Tensor} & 49.8 & 79.4  & 79.7 & 0.610 & 0.671 \\
 LMF \cite{Liu2018Efficient} & 50.0 & 80.6  & 81.0 & 0.608 & 0.677 \\
 MFN \cite{Zadeh2018Memory} & 49.1 & 79.6  & 80.6 & 0.618 &0.670 \\
 RAVEN \cite{RAVEN} & 50.2 & 79.0  & 79.4 & 0.605 & 0.680 \\
 MULT \cite{MULT} & 48.2 & 80.2  & 80.5 & 0.638 & 0.659 \\
 IMR \cite{MRM} & 48.7 & 80.6  & 81.0 & - & - \\
 QMF \cite{Quantum} & 47.9 & 80.7  & 79.8 & 0.640 & 0.658 \\
 MAG-BERT \cite{MAG-BERT} & 51.9 & 85.0 & 85.0 & 0.602 & 0.778 \\
 \hline
 $M^3SA$ (Hard) & \textbf{52.7} & \textbf{85.6}  & \textbf{85.5} & 0.587 & \textbf{0.789} \\
$M^3SA$ (Soft) & 52.5 & 85.2  & 85.1 & 0.599 & 0.781 \\
 \hline
 \end{tabular}}
  \vspace{-0.3cm}
\end{table}%

\subsection{Experimental Setting}
We use the CMU-MOSI  \cite{Zadeh2016MOSI} and CMU-MOSEI  \cite{MOSEI} datasets to evaluate the model. We provide details about the datasets, evaluation protocols, baseline methods, and other experimental details in Appendix.

During the training stage, we first update individual unimodal sub-networks with the modulated unimodal losses, after which the whole model is updated with the multimodal loss derived from MFM.

\subsection{Experimental Results}

\subsubsection{\textbf{Comparison with Baselines}}

In this section, we compare our proposed model with other baselines on two datasets CMU-MOSI \cite{Zadeh2016Multimodal} and CMU-MOSEI \cite{MOSEI}. As shown in Table~\ref{t1} and ~\ref{t2}, although \textbf{MAG-BERT} outperforms other existing methods and sets up a high baseline due to the effectiveness of BERT \cite{BERT}, it can be seen that both of our proposed $M^3SA$ (Hard) and $M^3SA$ (Soft) significantly outperform all baselines in most cases. Specifically, on CMU-MOSI dataset, our method achieves the best results on all metrics, and $M^3SA$ (Soft) outperforms MAG-BERT by 3.5\% on Acc7, 2.2\% on Acc2 and 2.1\% on F1 score. On CMU-MOSEI dataset, our proposed $M^3SA$ (Hard) yields 0.8\% improvement on Acc7, and 0.6\% on Acc2 and 0.5\% on F1 score compared with MAG-BERT. These results demonstrate the superiority of our proposed model, indicating the effectiveness of reaching optimal unimodal network and filtering out noisy modalities.



\subsubsection{\textbf{Ablation Study}}

In this section, we perform ablation studies to verify the effectiveness of each component by removing it from the model.

Aiming to verify the effectiveness of the designed modulation loss, we conduct experiments where \textbf{modulation loss} is removed (see the cases of `$M^3SA$ (Hard) (W/O ML)' and `$M^3SA$ (Soft) (W/O ML)' in Table~\ref{t3}). From the experimental results, it can be seen that removing the modulation loss degrades the performance of the model. Specifically, performance on Acc7, Acc2 and F1 score has seen a great drop. It is obvious that our proposed contrastive learning method is effective and can greatly boost the performance.

Meanwhile, we design two ablation experiments to investigate the contribution of MFM (see the cases of `$M^3SA$ (Hard) (W/O MFM)' and `$M^3SA$ (Soft) (W/O MFM) in Table~\ref{t3}). We can observe that without MFM, our model sees a greater drop in performance, which may be due to the reason that noisy information interferes the learning of other useful modalities. The results suggest the necessity to identify and filter out noisy modalities for a correct multimodal embedding, and in this way informative modalities can also be highlighted.

We also perform ablation study on the design of considering \textbf{baseline embedding} in MFM (see the cases of `$M^3SA$ (Hard) (W/O BE)' and `$M^3SA$ (Soft) (W/O BE) in Table~\ref{t3}). We can see from the results that removing the compensation of baseline embedding in MFM degrades the performance of $M^3SA$ severely compared to other cases. Specifically, the performance drops even greater than the cases W/O MFM. It may be because, despite the removal of noisy information, modality-specific information of the noisy modality is lost. The results indicate that \textbf{the learning of baseline embedding in MFM is of necessity}, for it compensates the filtered modality-specific information.

\begin{table}[t]
\centering
 \caption{ \label{t3}\textbf{ Ablation studies on the CMU-MOSI dataset.} The `ML', `MFM' and `BE' refer to our proposed modulation loss, modality filter module and baseline embedding, respectively.}
\resizebox{.95\columnwidth}{!}{\begin{tabular}{c|c|c|c|c|c}
 \hline
    & Acc7 & Acc2 & F1 & MAE & Corr \\
 \hline
 $M^3SA$ (Hard) (W/O ML) & 44.9 & 84.2 & 84.2 & 0.743 & 0.786  \\
 $M^3SA$ (Soft) (W/O ML) & 46.2 & 85.0 & 84.9 & 0.729 & 0.794  \\
 $M^3SA$ (Hard) (W/O MFM) & \textbf{47.0} & 84.8 & 84.8 & 0.725 & 0.791  \\
 $M^3SA$ (Soft) (W/O MFM) & 44.2 & 83.9 & 83.9 & 0.737 & 0.794  \\
 $M^3SA$ (Hard) (W/O BE) & 46.1 & 84.2 & 84.2 & 0.728 & 0.788  \\
 $M^3SA$ (Soft) (W/O BE) & 46.1 & 83.9 & 83.9 & 0.733 & 0.794  \\
 \hline
 $M^3SA$ (Hard) & 45.5 & 85.3  & 85.3 & 0.730 & 0.793 \\
 $M^3SA$ (Soft) & 46.4 & \textbf{85.7}  & \textbf{85.6} & \textbf{0.714} & \textbf{0.794} \\


 \hline
 \end{tabular}}
\end{table}%

\begin{table}[t]
\centering
 \caption{ \label{t4}\textbf{ Discussion on the fusion strategies.} Graph fusion \cite{ARGF} regards each unimodal, bimodal, and trimodal interaction as one node, and explicitly models their relationship. Tensor fusion \cite{Zadeh2017Tensor} applies outer product to explore interactions, which introduces a large amount of parameters and has high space complexity. The defaulted fusion method is addition.
 }
\resizebox{.95\columnwidth}{!}{\begin{tabular}{c|c|c|c|c|c}
 \hline
    & Acc7 & Acc2 & F1 & MAE & Corr \\
 \hline
 Concatenation+FC (Hard) & \textbf{48.0} & 84.0 & 83.9 & 0.744  &   0.783 \\
 Addition (Hard)  & 45.5 & 85.3  & 85.3 & 0.730 & 0.793  \\
 Tensor Fusion (Hard)  & 43.1 & 84.3 & 84.3 & 0.772 & 0.786 \\
 Graph Fusion (Hard)  & 45.7 & 84.6 & 84.6 & 0.759 &  0.772  \\
  Concatenation+FC (Soft) & 45.4 & 84.4 & 84.4 & 0.740 &  0.790  \\
 Addition (Soft)  & 46.4 & \textbf{85.7}  & \textbf{85.6} & \textbf{0.714} & \textbf{0.794}  \\
 Tensor Fusion (Soft)  & 43.8 & 84.7 & 84.7 & 0.742 &    0.787\\
 Graph Fusion (Soft)  & 46.6 & 84.7 & 84.6 & 0.748 &  0.775\\
 \hline
 \end{tabular}}
\end{table}%


\begin{figure}
\setlength{\abovecaptionskip}{0.05cm}
\setlength{\belowcaptionskip}{-0.3cm}
\centering
\includegraphics[scale=0.2]{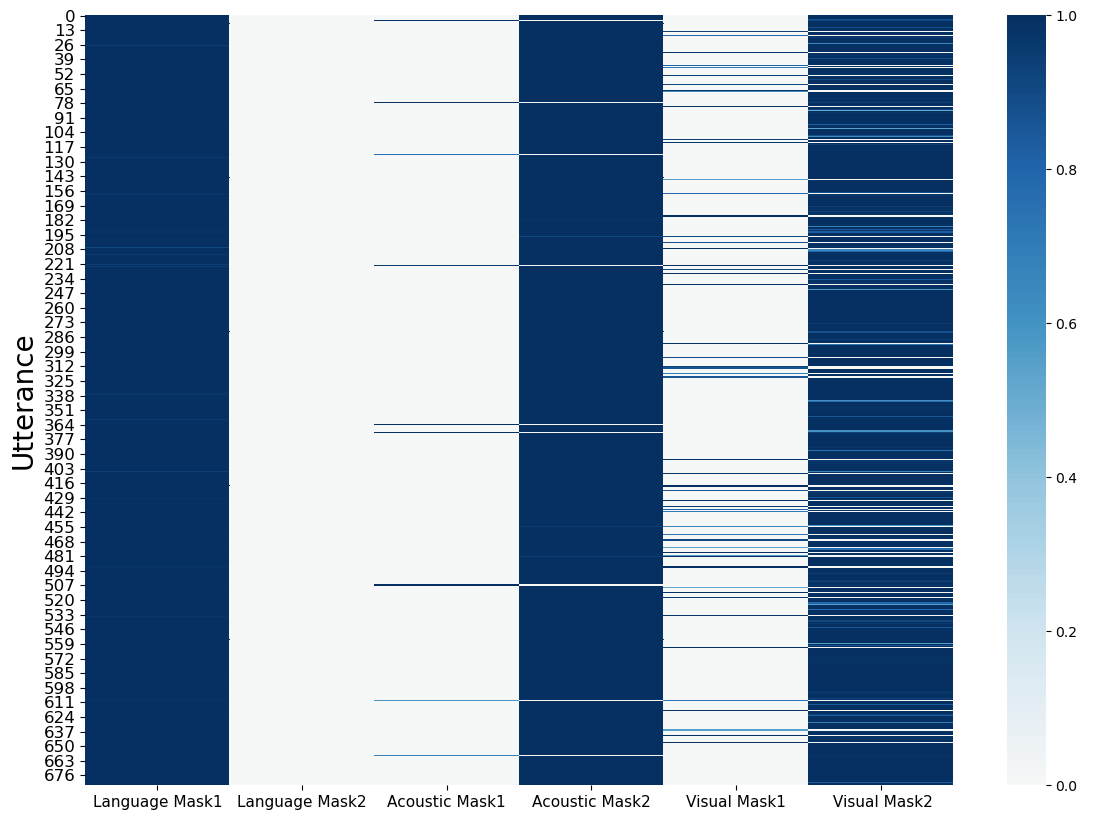}
\caption{\label{tsne}\textbf{Visualization of the Mask Values of the Three Modality Learned by Soft Filter.}
}
\end{figure}

\subsubsection{\textbf{Analysis of Generalization Ability}}
We also conduct experiments to verify that our proposed $M^3SA$ is generalized to be applied with different fusion strategies. Previous work mostly rely on sophisticated fusion methods to sufficiently learn cross-modal dynamics to reach satisfactory results. Unlike them, our proposed model can achieve state-of-the-art performance with simple fusion strategies. As shown in Table~\ref{t4}, \textbf{even with simple and direct fusion methods like concatenation and element-wise addition of unimodal representations, $M^3SA$ still outperforms all baselines in most cases}. Note that despite the choice of $M^3SA$ (Hard) or $M^3SA$ (Soft), \textbf{all the variants of our model reach the state-of-the-art performance compared to baselines}. A conclusion can be reached that our designed modulation loss and MFM is effective and of satisfactory generalization ability. Also note that our proposed modulation loss and MFM can be applied to any cross-modal scenarios.


As shown in the Table, combining all the evaluation metrics, the simple fusion method, i.e., Addition performs best. We argue that apart from the modulation loss which can help to learn better unimodal representation, it is partly because we use the same classifier $C$ to regularize the feature distributions of unimodal and multimodal representations which forces them to have the same distribution, such that direct addition is strong enough to explore the complementary information and interactions between modalities.  Instead, the high-complex learnable fusion methods may introduce noise to the distribution, which degrades the performance. Specifically, we can observe that tenser fusion \cite{Zadeh2017Tensor} gets a relatively unfavorable results. The reason for it could be that tensor fusion implements the outer product on vectors of all modalities, which may change the distribution of high-level features and exhaust the deep network for introducing a lot of computation and parameters. 


\subsubsection{\textbf{Analysis on the Modality Importance}}
We provide a visualization for the learned mask value of the soft filter for the testing utterances, aiming to verify the effectiveness of MFM to identify and filter out noisy modalities.
Note that the value of ‘Mask1' and ’Mask2' represents the percentage of the preserved information and filtered information of the corresponding modality. We can infer from Fig.~\ref{tsne} that, the language modality is the most informative modality that is rarely filtered out (and this conclusion is consistent with other works \cite{TASLP}). Contrary to it, the acoustic modality is frequently identified as noisy and filtered out which is the most uninformative modality. It can be seen that our MFM is capable to identify and filter out noisy modalities, which can also highlight the role of informative modalities when noisy information is filtered. Notably, the mean mask value is 0.998, 0.012, 0.088 for language, acoustic, and visual modalities, respectively.

Also, from the visualization results we can observe that the learned mask value approximates the 0-1 distribution (i.e, a modality is identified as either very informative or very noisy), which differs from existing attention mechanisms and the difference is mostly due to our defined scale factor $\lambda$ and penalty loss $l^p$. Apart from highlighting important modalities as in attention mechanisms, our MFM can reach better filtering effect and can be integrated with any fusion mechanisms.
The visualization of $M^3SA$ (Hard) is similar, which is not presented due to the page limitations.

\section{Conclusions}
We propose novel MSA framework to modulate the learning of unimodal and cross-modal dynamics, which is capable of exploring an optimal solution for unimodal networks and filtering out noisy modalities. Specifically, modulation loss can  modulate the learning of unimodal networks based on their confidence of prediction, while modality filter module can filter out noisy modalities for a correct multimodal embedding. 
Experiments demonstrate that our model outperforms state-of-the-art methods in two datasets.

\bibliography{anthology,custom}

\begin{thebibliography}{45}
\expandafter\ifx\csname natexlab\endcsname\relax\def\natexlab#1{#1}\fi

\bibitem[{Akhtar et~al.(2019)Akhtar, Chauhan, Ghosal, Poria, Ekbal, and
  Bhattacharyya}]{MTL}
Md~Shad Akhtar, Dushyant Chauhan, Deepanway Ghosal, Soujanya Poria, Asif Ekbal,
  and Pushpak Bhattacharyya. 2019.
\newblock Multi-task learning for multi-modal emotion recognition and sentiment
  analysis.
\newblock In \emph{Proceedings of the 2019 Conference of the North American
  Chapter of the Association for Computational Linguistics: Human Language
  Technologies, Volume 1 (Long and Short Papers)}, pages 370--379.

\bibitem[{Chauhan et~al.(2019)Chauhan, Akhtar, Ekbal, and Bhattacharyya}]{CIA}
Dushyant~Singh Chauhan, Md~Shad Akhtar, Asif Ekbal, and Pushpak Bhattacharyya.
  2019.
\newblock Context-aware interactive attention for multi-modal sentiment and
  emotion analysis.
\newblock In \emph{Proceedings of the 2019 Conference on Empirical Methods in
  Natural Language Processing and the 9th International Joint Conference on
  Natural Language Processing (EMNLP-IJCNLP)}, pages 5651--5661.

\bibitem[{Chen et~al.(2017)Chen, Wang, Liang, Baltrus$\check{a}$itis, Zadeh,
  and Morency}]{Chen2018Multimodal}
Minghai Chen, Sen Wang, Paul~Pu Liang, Tadas Baltrus$\check{a}$itis, Amir
  Zadeh, and Louis~Philippe Morency. 2017.
\newblock Multimodal sentiment analysis with word-level fusion and
  reinforcement learning.
\newblock In \emph{19th ACM International Conference on Multimodal Interaction
  (ICMI'17)}, pages 163--171.

\bibitem[{Degottex et~al.(2014)Degottex, Kane, Drugman, Raitio, and
  Scherer}]{Degottex2014COVAREP}
Gilles Degottex, John Kane, Thomas Drugman, Tuomo Raitio, and Stefan Scherer.
  2014.
\newblock Covarep: A collaborative voice analysis repository for speech
  technologies.
\newblock In \emph{ICASSP}, pages 960--964.

\bibitem[{Devlin et~al.(2019)Devlin, Chang, Lee, and Toutanova}]{BERT}
Jacob Devlin, Ming-Wei Chang, Kenton Lee, and Kristina Toutanova. 2019.
\newblock \href {https://doi.org/10.18653/v1/N19-1423} {{BERT}: Pre-training of
  deep bidirectional transformers for language understanding}.
\newblock In \emph{Proceedings of the 2019 Conference of the North {A}merican
  Chapter of the Association for Computational Linguistics: Human Language
  Technologies, Volume 1 (Long and Short Papers)}, pages 4171--4186,
  Minneapolis, Minnesota. Association for Computational Linguistics.

\bibitem[{George and Marcel(2021)}]{focal}
Anjith George and S{\'e}bastien Marcel. 2021.
\newblock Cross modal focal loss for rgbd face anti-spoofing.
\newblock In \emph{Proceedings of the IEEE/CVF Conference on Computer Vision
  and Pattern Recognition}, pages 7882--7891.

\bibitem[{Gkoumas et~al.(2021)Gkoumas, Li, Lioma, Yu, and wei
  Song}]{Gkoumas2021WhatMT}
Dimitris Gkoumas, Qiuchi Li, C.~Lioma, Yijun Yu, and Da~wei Song. 2021.
\newblock What makes the difference? an empirical comparison of fusion
  strategies for multimodal language analysis.
\newblock \emph{Information Fusion}, 66:184--197.

\bibitem[{Hazarika et~al.(2020)Hazarika, Zimmermann, and Poria}]{MISA}
Devamanyu Hazarika, R.~Zimmermann, and Soujanya Poria. 2020.
\newblock Misa: Modality-invariant and -specific representations for multimodal
  sentiment analysis.
\newblock \emph{Proceedings of the 28th ACM International Conference on
  Multimedia}.

\bibitem[{Hou et~al.(2019)Hou, Tang, Zhang, Kong, and Zhao}]{HPFN}
Ming Hou, Jiajia Tang, Jianhai Zhang, Wanzeng Kong, and Qibin Zhao. 2019.
\newblock Deep multimodal multilinear fusion with high-order polynomial
  pooling.
\newblock In \emph{Advances in Neural Information Processing Systems}, pages
  12113--12122.

\bibitem[{iMotions 2017(2017)}]{FACET}
iMotions 2017. 2017.
\newblock imotions.
\newblock \emph{Facial expression analysis}.

\bibitem[{Kingma and Ba(2015)}]{Kingma2014Adam}
Diederik~P Kingma and Jimmy Ba. 2015.
\newblock Adam: A method for stochastic optimization.
\newblock In \emph{Proceedings of International Conference on Learning
  Representations (ICLR)}.

\bibitem[{Kingma and Welling(2013)}]{hard2}
Diederik~P Kingma and Max Welling. 2013.
\newblock Auto-encoding variational bayes.
\newblock \emph{arXiv preprint arXiv:1312.6114}.

\bibitem[{Li et~al.(2021)Li, Gkoumas, Lioma, and Melucci}]{Quantum}
Qiuchi Li, Dimitris Gkoumas, Christina Lioma, and Massimo Melucci. 2021.
\newblock \href {https://doi.org/https://doi.org/10.1016/j.inffus.2020.08.006}
  {Quantum-inspired multimodal fusion for video sentiment analysis}.
\newblock \emph{Information Fusion}, 65:58 -- 71.

\bibitem[{Liang et~al.(2019)Liang, Lim, Tsai, Salakhutdinov, and Morency}]{MMB}
Paul~Pu Liang, Yao~Chong Lim, Y.~H. Tsai, Ruslan~R. Salakhutdinov, and
  Louis-Philippe Morency. 2019.
\newblock Strong and simple baselines for multimodal utterance embeddings.
\newblock In \emph{NAACL}, pages 2599--2609.

\bibitem[{Liang et~al.(2018)Liang, Liu, Zadeh, and Morency}]{RMFN}
Paul~Pu Liang, Ziyin Liu, Amir Zadeh, and Louis~Philippe Morency. 2018.
\newblock Multimodal language analysis with recurrent multistage fusion.
\newblock In \emph{EMNLP}, pages 150--161.

\bibitem[{Liu et~al.(2018)Liu, Shen, Liang, Zadeh, and
  Morency}]{Liu2018Efficient}
Zhun Liu, Ying Shen, Paul~Pu Liang, Amir Zadeh, and Louis~Philippe Morency.
  2018.
\newblock Efficient low-rank multimodal fusion with modality-specific factors.
\newblock In \emph{ACL}, pages 2247--2256.

\bibitem[{Louizos et~al.(2017)Louizos, Welling, and Kingma}]{hard1}
Christos Louizos, Max Welling, and Diederik~P Kingma. 2017.
\newblock Learning sparse neural networks through $ l\_0 $ regularization.
\newblock \emph{arXiv preprint arXiv:1712.01312}.

\bibitem[{Mai et~al.(2019)Mai, Hu, and Xing}]{HFFN}
Sijie Mai, Haifeng Hu, and Songlong Xing. 2019.
\newblock Divide, conquer and combine: Hierarchical feature fusion network with
  local and global perspectives for multimodal affective computing.
\newblock In \emph{ACL}.

\bibitem[{Mai et~al.(2020{\natexlab{a}})Mai, Hu, and Xing}]{ARGF}
Sijie Mai, Haifeng Hu, and Songlong Xing. 2020{\natexlab{a}}.
\newblock Modality to modality translation: An adversarial representation
  learning and graph fusion network for multimodal fusion.
\newblock In \emph{Proceedings of the AAAI Conference on Artificial
  Intelligence}, volume~34, pages 164--172.

\bibitem[{Mai et~al.(2021{\natexlab{a}})Mai, Hu, and Xing}]{mai2021unimodal}
Sijie Mai, Haifeng Hu, and Songlong Xing. 2021{\natexlab{a}}.
\newblock A unimodal representation learning and recurrent decomposition fusion
  structure for utterance-level multimodal embedding learning.
\newblock \emph{IEEE Transactions on Multimedia}.

\bibitem[{Mai et~al.(2020{\natexlab{b}})Mai, Xing, He, Zeng, and
  Hu}]{mai2020analyzing}
Sijie Mai, Songlong Xing, Jiaxuan He, Ying Zeng, and Haifeng Hu.
  2020{\natexlab{b}}.
\newblock \href {http://arxiv.org/abs/2011.13572} {Analyzing unaligned
  multimodal sequence via graph convolution and graph pooling fusion}.

\bibitem[{Mai et~al.(2021{\natexlab{b}})Mai, Xing, and Hu}]{TASLP}
Sijie Mai, Songlong Xing, and Haifeng Hu. 2021{\natexlab{b}}.
\newblock \href {https://doi.org/10.1109/TASLP.2021.3068598} {Analyzing
  multimodal sentiment via acoustic- and visual-lstm with channel-aware
  temporal convolution network}.
\newblock \emph{IEEE/ACM Transactions on Audio, Speech, and Language
  Processing}, 29:1424--1437.

\bibitem[{Mnih and Gregor(2014)}]{hard3}
Andriy Mnih and Karol Gregor. 2014.
\newblock Neural variational inference and learning in belief networks.
\newblock In \emph{International Conference on Machine Learning}, pages
  1791--1799. PMLR.

\bibitem[{Olson(1977)}]{Olson1977From}
David Olson. 1977.
\newblock From utterance to text: The bias of language in speech and writing.
\newblock \emph{Harvard Educational Review}, 47(3):257--281.

\bibitem[{Pham et~al.(2019)Pham, Liang, Manzini, Morency, and
  Barnab\v{a}s}]{MCTN}
Hai Pham, Paul~Pu Liang, Thomas Manzini, Louis~Philippe Morency, and Pocz\v{o}s
  Barnab\v{a}s. 2019.
\newblock Found in translation: Learning robust joint representations by cyclic
  translations between modalities.
\newblock In \emph{AAAI}, pages 6892--6899.

\bibitem[{Poria et~al.(2017{\natexlab{a}})Poria, Cambria, Bajpai, and
  Hussain}]{Poria2017A}
Soujanya Poria, Erik Cambria, Rajiv Bajpai, and Amir Hussain.
  2017{\natexlab{a}}.
\newblock A review of affective computing: From unimodal analysis to multimodal
  fusion.
\newblock \emph{Information Fusion}, 37:98--125.

\bibitem[{Poria et~al.(2017{\natexlab{b}})Poria, Cambria, Hazarika, Majumder,
  Zadeh, and Morency}]{Poria2017Context}
Soujanya Poria, Erik Cambria, Devamanyu Hazarika, Navonil Majumder, Amir Zadeh,
  and Louis~Philippe Morency. 2017{\natexlab{b}}.
\newblock Context-dependent sentiment analysis in user-generated videos.
\newblock In \emph{ACL}, pages 873--883.

\bibitem[{Poria et~al.(2016)Poria, Chaturvedi, Cambria, and
  Hussain}]{Poria2017Convolutional}
Soujanya Poria, Iti Chaturvedi, Erik Cambria, and Amir Hussain. 2016.
\newblock Convolutional mkl based multimodal emotion recognition and sentiment
  analysis.
\newblock In \emph{Proceedings of IEEE International Conference on Data Mining
  (ICDM)}, pages 439--448.

\bibitem[{Rahman et~al.(2020)Rahman, Hasan, Lee, Zadeh, Mao, Morency, and
  Hoque}]{MAG-BERT}
Wasifur Rahman, M.~Hasan, Sangwu Lee, Amir Zadeh, Chengfeng Mao, Louis-Philippe
  Morency, and E.~Hoque. 2020.
\newblock Integrating multimodal information in large pretrained transformers.
\newblock \emph{Proceedings of the conference. Association for Computational
  Linguistics. Meeting}, 2020:2359--2369.

\bibitem[{Rozgic et~al.(2012)Rozgic, Ananthakrishnan, Saleem, Kumar, and
  Prasad}]{Rozgic2012Ensemble}
V.~Rozgic, S.~Ananthakrishnan, S.~Saleem, R.~Kumar, and R.~Prasad. 2012.
\newblock Ensemble of svm trees for multimodal emotion recognition.
\newblock In \emph{Signal and Information Processing Association Summit and
  Conference}, pages 1--4.

\bibitem[{Schlichtkrull et~al.(2020)Schlichtkrull, De~Cao, and Titov}]{mask}
Michael~Sejr Schlichtkrull, Nicola De~Cao, and Ivan Titov. 2020.
\newblock Interpreting graph neural networks for nlp with differentiable edge
  masking.
\newblock \emph{arXiv preprint arXiv:2010.00577}.

\bibitem[{Shenoy and Sardana(2020)}]{Multilogue-Net}
Aman Shenoy and Ashish Sardana. 2020.
\newblock Multilogue-net: A context aware rnn for multi-modal emotion detection
  and sentiment analysis in conversation.
\newblock \emph{arXiv preprint arXiv:2002.08267}.

\bibitem[{Tsai et~al.(2019{\natexlab{a}})Tsai, Bai, Liang, Kolter, Morency, and
  Salakhutdinov}]{MULT}
Yao-Hung~Hubert Tsai, Shaojie Bai, Paul~Pu Liang, J.~Zico Kolter,
  Louis-Philippe Morency, and Ruslan Salakhutdinov. 2019{\natexlab{a}}.
\newblock Multimodal transformer for unaligned multimodal language sequences.
\newblock In \emph{ACL}.

\bibitem[{Tsai et~al.(2019{\natexlab{b}})Tsai, Liang, Zadeh, Morency, and
  Salakhutdinov}]{MFM}
Yao Hung~Hubert Tsai, Paul~Pu Liang, Amir Zadeh, Louis~Philippe Morency, and
  Ruslan Salakhutdinov. 2019{\natexlab{b}}.
\newblock Learning factorized multimodal representations.
\newblock In \emph{ICLR}.

\bibitem[{Tsai et~al.(2020)Tsai, Ma, Yang, Salakhutdinov, and Morency}]{MRM}
Yao-Hung~Hubert Tsai, Martin~Q. Ma, Muqiao Yang, Ruslan Salakhutdinov, and
  Louis-Philippe Morency. 2020.
\newblock \href {http://arxiv.org/abs/arXiv:2004.14198} {Multimodal routing:
  Improving local and global interpretability of multimodal language analysis}.
\newblock \emph{arXiv preprint arXiv:2001.08735, 2020}.

\bibitem[{Vaswani et~al.(2017)Vaswani, Shazeer, Parmar, Uszkoreit, Jones,
  Gomez, Kaiser, and Polosukhin}]{transformer}
Ashish Vaswani, Noam Shazeer, Niki Parmar, Jakob Uszkoreit, Llion Jones,
  Aidan~N Gomez, {\L}ukasz Kaiser, and Illia Polosukhin. 2017.
\newblock Attention is all you need.
\newblock In \emph{NIPS}, pages 5998--6008.

\bibitem[{Wang et~al.(2019)Wang, Shen, Liu, Liang, Zadeh, and Morency}]{RAVEN}
Yansen Wang, Ying Shen, Zhun Liu, Paul~Pu Liang, Amir Zadeh, and Louis-Philippe
  Morency. 2019.
\newblock Words can shift: Dynamically adjusting word representations using
  nonverbal behaviors.
\newblock In \emph{AAAI}, volume~33, pages 7216--7223.

\bibitem[{Wollmer et~al.(2013)Wollmer, Weninger, Knaup, Schuller, Sun, Sagae,
  and Morency}]{Wollmer2013YouTube}
Martin Wollmer, Felix Weninger, Tobias Knaup, Bjorn Schuller, Congkai Sun,
  Kenji Sagae, and Louis~Philippe Morency. 2013.
\newblock Youtube movie reviews: Sentiment analysis in an audio-visual context.
\newblock \emph{IEEE Intelligent Systems}, 28(3):46--53.

\bibitem[{Yang et~al.(2020)Yang, Xu, and Gao}]{CM-BERT}
Kaicheng Yang, Hua Xu, and Kai Gao. 2020.
\newblock Cm-bert: Cross-modal bert for text-audio sentiment analysis.
\newblock In \emph{Proceedings of the 28th ACM International Conference on
  Multimedia}, pages 521--528.

\bibitem[{Zadeh et~al.(2017)Zadeh, Chen, Poria, Cambria, and
  Morency}]{Zadeh2017Tensor}
Amir Zadeh, Minghai Chen, Soujanya Poria, Erik Cambria, and Louis~Philippe
  Morency. 2017.
\newblock Tensor fusion network for multimodal sentiment analysis.
\newblock In \emph{EMNLP}, pages 1114--1125.

\bibitem[{Zadeh et~al.(2018{\natexlab{a}})Zadeh, Liang, Mazumder, Poria,
  Cambria, and Morency}]{Zadeh2018Memory}
Amir Zadeh, Paul~Pu Liang, Navonil Mazumder, Soujanya Poria, Erik Cambria, and
  Louis~Philippe Morency. 2018{\natexlab{a}}.
\newblock Memory fusion network for multi-view sequential learning.
\newblock In \emph{AAAI}, pages 5634--5641.

\bibitem[{Zadeh et~al.(2018{\natexlab{b}})Zadeh, Liang, Vanbriesen, Poria,
  Tong, Cambria, Chen, and Morency}]{MOSEI}
Amir Zadeh, Paul~Pu Liang, Jonathan Vanbriesen, Soujanya Poria, Edmund Tong,
  Erik Cambria, Minghai Chen, and Louis~Philippe Morency. 2018{\natexlab{b}}.
\newblock Multimodal language analysis in the wild: Cmu-mosei dataset and
  interpretable dynamic fusion graph.
\newblock In \emph{ACL}, pages 2236--2246.

\bibitem[{Zadeh et~al.(2016{\natexlab{a}})Zadeh, Zellers, Pincus, and
  Morency}]{Zadeh2016MOSI}
Amir Zadeh, Rowan Zellers, Eli Pincus, and Louis~Philippe Morency.
  2016{\natexlab{a}}.
\newblock Mosi: Multimodal corpus of sentiment intensity and subjectivity
  analysis in online opinion videos.
\newblock \emph{IEEE Intelligent Systems}, 31(6):82--88.

\bibitem[{Zadeh et~al.(2016{\natexlab{b}})Zadeh, Zellers, Pincus, and
  Morency}]{Zadeh2016Multimodal}
Amir Zadeh, Rowan Zellers, Eli Pincus, and Louis~Philippe Morency.
  2016{\natexlab{b}}.
\newblock Multimodal sentiment intensity analysis in videos: Facial gestures
  and verbal messages.
\newblock \emph{IEEE Intelligent Systems}, 31(6):82--88.

\bibitem[{Zhang et~al.(2019)Zhang, Li, Zhu, and Zhou}]{rl2}
Dong Zhang, Shoushan Li, Qiaoming Zhu, and Guodong Zhou. 2019.
\newblock Effective sentiment-relevant word selection for multi-modal sentiment
  analysis in spoken language.
\newblock In \emph{Proceedings of the 27th ACM International Conference on
  Multimedia}, pages 148--156.

\end{thebibliography}
\bibliographystyle{acl_natbib}




\appendix
\section*{Appendix}
\section{\textbf{Unimodal Network: $\bm{F}^m$}}
\label{sec:unimodal}
Since Transformer-based \cite{transformer} structure enables parallel computation in time dimension and can learn longer temporal dependency in long  sequences, we apply Transformer-based \cite{transformer} architectures to build up the unimodal learning networks. Specifically, for acoustic and visual modalities, we apply the standard Transformer to extract the high-level unimodal representations. For  language modality, the large-pretrained Transformer model, i.e., BERT \cite{BERT} is applied to extract the language representation. The equations are shown as below:
\begin{equation}
\setlength{\abovedisplayskip}{3pt}
\setlength{\belowdisplayskip}{3pt}
\label{eq6}
\begin{split}
   \bm{\hat{X}}^{l}&=\text{BERT}(\bm{U}^l)\\
    \bm{X}^{l}&=\operatorname{Conv} 1 \mathrm{D}\left( \bm{\hat{X}}^{l}, K_l\right) \in \mathbb{R}^{T_l \times d}\\
    \bm{x}^{l}&= \bm{X}^{l}_{T_l}\in \mathbb{R}^{d}\\
\end{split}
\end{equation}
where $\operatorname{Conv} 1 \mathrm{D}$ denotes the temporal convolution operation with $K_l$ being the kernel size, which is used for mapping the output dimensionality of BERT to the shared dimensionality $d$ that are equal for all modalities. Note that $\bm{x}^{l}$ is the feature embedding of $\bm{X}^{l}$ in the last time step, and we only use the feature embedding of the last time step to conduct fusion and prediction such that our model is suitable for handling the fusion of unimodal sequences of various length. For acoustic and visual modalities, the equations are presented as follows:
\begin{equation}
\setlength{\abovedisplayskip}{3pt}
\setlength{\belowdisplayskip}{3pt}
\label{eq6}
\begin{split}
\bm{\hat{X}}^{m}&=\operatorname{Conv} 1 \mathrm{D}\left( \bm{U}^{m}, K_m\right) \in \mathbb{R}^{T_m \times d}\\
   \bm{X}^{m}&=\text{Transformer}(\bm{\hat{X}}^{m}) \in \mathbb{R}^{T_m \times d}\\
    \bm{x}^{m}&= \bm{X}^{m}_{T_m}\in \mathbb{R}^{d}, \ m\in \{a,v \}\\
\end{split}
\end{equation}
Different from the language processing procedure, the temporal convolution operation for the other modalities is used before the Transformer to map the feature dimensionality to the same one.

\section{\textbf{Multimodal Network: $\bm{F}^M$}}
\label{sec:multimodal}
Our algorithm is independent of the concrete fusion mechanism, and we can inject various fusion methods into our multimodal learning structure. In this paper, we mainly investigate four fusion methods to verify the effectiveness of our algorithm. Note that since the unimodal and multimodal representations share the same classifier $C$, the dimensionality of the fused multimodal representation shall be the same as that of the unimodal representations. The fusion methods are illustrated as follows:

1) \textbf{Direct Addition}:
\begin{equation}
 \bm{x^M} = \bm{x}^l + \bm{x}^a + \bm{x}^v
\end{equation}
where $\bm{x^M}\in \mathbb{R}^{d} $ is the multimodal representation. Since the addition will not change the feature dimensionality, we need not to apply a learnable layer such as fully-connected layer to change the feature dimensionality of the multimodal representation. Therefore, this method of fusion is learnable. In our experiment, we show that even with such a simple fusion method, our algorithm can still reach very competitive performance.

2) \textbf{Concatenation}:
\begin{equation}
 \bm{x^M} = FC(\bm{x}^l\oplus\bm{x}^a\oplus\bm{x}^v)
\end{equation}
where $FC\in \mathbb{R}^{3\times d}\rightarrow \mathbb{R}^{d}$ denotes fully-connected network to map the feature dimensionality to $d$. This method is learnable as it uses fully-connected layers to inject the multimodal representation into the common embedding space as that of the unimodal representations. Together with Direct Addition, it serves as the baseline fusion methods throughout the researches of multimodal learning.

3) \textbf{Tensor Fusion}: Tensor fusion \cite{Zadeh2017Tensor} is a widely-used fusion algorithm that attracts significant attention \cite{HFFN,Liu2018Efficient,HPFN}. By applying outer product over the unimodal representations, the generated multimodal representation has the highest expressive power but meanwhile is high-dimensional. The equations for tensor fusion are shown below:
\begin{equation}
\label{eqt}
\setlength{\abovedisplayskip}{3pt}
\setlength{\belowdisplayskip}{3pt}
  \bm{x}^{m'}=[\bm{x}^{m},\ 1],\ \ m \in \{l,v,a \}
\end{equation}
\begin{equation}
\setlength{\abovedisplayskip}{3pt}
\setlength{\belowdisplayskip}{3pt}
  \bm{\hat{x^M}}= FC(\bigotimes_{m}\bm{x}^{m'}), \ \ \bm{x}^{m'}\in \mathbb{R}^{d+1}
\end{equation}
where $\bigotimes$ denotes outer product of a set of vectors, $FC\in \mathbb{R}^{(d+1)^3}\rightarrow \mathbb{R}^{d}$ denotes fully-connected network to map the feature dimensionality to $d$. In Eq.~\ref{eqt}, each unimodal representation is padded with \emph{1s} to retain interactions of any subset of modalities as in \cite{Zadeh2017Tensor}.

4) \textbf{Graph Fusion}: Graph fusion \cite{ARGF} regards each modality as one node, and conduct message passing between nodes to explore unimodal, bimodal, and trimodal dynamics. The final graph representation is obtained by averaging the node embedding. For more details, please refer to the Graph Fusion Network in \cite{ARGF}.

\section{Experimental Setting}
\label{sec:experimental setting}

\subsection{Datasets}

In this paper, two of the most commonly used public datasets, i.e, CMU-MOSEI \cite{MOSEI} and CMU-MOSI \cite{Zadeh2016MOSI} are adopted to perform MSA in our experiments:

1) \textbf{CMU-MOSI} is a widely-used dataset for multimodal sentiment analysis, which is a collection of 2199 opinion video clips. Each opinion video is annotated with sentiment on a [-3,3]. 
To be consistent with prior works, we use 1,284  utterances for training, 229 utterances for validation, and 686 utterances for testing.

2) \textbf{CMU-MOSEI} is a large dataset of multimodal sentiment analysis and emotion recognition. The dataset consists of 23454 video utterances from more than 1000 YouTube speakers, covering 250 distinct topics. All the sentences utterance are randomly chosen from various topics and monologue videos, and each utterance is annotated on two views: emotion of six different values, and sentiment in the range [-3,3]. In our work, we use the sentiment label to perform MSA. We use 16,265 utterances as training set, 1,869 utterances as validation set, and 4,643 utterances as testing set.

\subsection{Evaluation Protocol}
In our experiments, the evaluation metrics for CMU-MOSEI are the same as those for CMU-MOSI dataset. We adopt various metrics to evaluate the performance of each model: 1) Acc7: 7-way accuracy, sentiment score classification; 2) Acc2: binary accuracy, positive or negative; 3) F1 score; 4) MAE: mean absolute error and 5) Corr: the correlation of the model's prediction.



\subsection{Baselines}

We compare our proposed model with the following state-of-the-art models:

1) \textbf{Early Fusion LSTM} (\textbf{EF-LSTM}), which is the baseline fusion approach that concatenates the input features of different modalities at word-level, and then sends the concatenated features to an LSTM layer. EF-LSTM is an RNN-based word-level fusion model.

2) \textbf{Late Fusion LSTM} (\textbf{LF-LSTM}), which is another baseline method that uses an LSTM network for each modality to extract unimodal features and infer decision, and then combine the unimodal decisions by voting mechanism, etc.

3) \textbf{Recurrent Attended Variation Embedding Network} (\textbf{RAVEN}) \cite{RAVEN}, which models human language by shifting word representations based on the features of the facial expressions and vocal patterns. It is an RNN-based word-level fusion approaches.

4) \textbf{Memory Fusion Network} (\textbf{MFN}) \cite{Zadeh2018Memory} is also an RNN-based word-level fusion method, which includes three components. The first component is the systems of LSTMs which is used to model unimodal dynamics. The latter components are delta-attention module and multi-view gated memory network  which are used for discovering cross-modal dynamics through time.

5) \textbf{Multimodal Transformer} (\textbf{MULT}) \cite{MULT}, which  learns joint multimodal representation by translating source modality into target modality. It is a transformer-based model.

6) \textbf{Interpretable Modality Fusion} (\textbf{IMR}) \cite{MRM}, which improves the interpretable  ability of MULT by introducing the multimodal routing mechanism. IMR is also a transformer-based model.

7) \textbf{Tensor Fusion Network} (\textbf{TFN}) \cite{Zadeh2017Tensor}, which applies 3-fold outer product from modality embeddings to jointly learn unimodal, bimodal and trimodal interactions.

8) \textbf{Low-rank Modality Fusion} (\textbf{LMF}) \cite{Liu2018Efficient}, which leverages low-rank
weight tensors to reduce the complexity of tensor fusion without compromising on performance.

9) \textbf{Quantum-inspired Multimodal Fusion} (\textbf{QMF}) \cite{Quantum}, which addresses the interpretable problem of multimodal fusion by taking inspiration from the quantum theory.

10) \textbf{Multimodal Adaption Gate BERT} (\textbf{MAG-BERT}) \cite{MAG-BERT}: MAG-BERT proposes an attachment to BERT and XLNet called Multimodal Adaptation Gate (MAG), which allows BERT and XLNet to accept multimodal nonverbal data during fine-tuning. The feature extraction method of MAG-BERT is the same as that of our method, which ensures fair comparison. MAG-BERT is currently the state-of-the-art algorithm on multimodal sentiment analysis.


\subsection{Experimental Details}
For each baseline (except for QMF \cite{Quantum} whose codes are unavailable), following \cite{Gkoumas2021WhatMT}, we first perform fifty-times random grid search on the hyper-parameters to fine-tune the model, and save the hyper-parameter setting that reaches the best performance. After that, we train each model with the best hyper-parameters setting for five times, and the final results are obtained by calculating the mean results.

For CMU-MOSEI dataset, the input dimensionality of language, audio, and visual modality is 768, 74, and 35, respectively. While for CMU-MOSI, the input dimensionality of language, audio, and visual modality is 768, 74, and 47, respectively. For feature extraction,  Facet \cite{FACET} \footnote{ iMotions 2017. https://imotions.com/} is used for the visual modality to extract a set of features that are composed of facial action units, facial landmarks, head pose, etc. These visual features are extracted from the video utterance at the frequency of 30Hz to form a sequence of facial gestures over time. COVAREP \cite{Degottex2014COVAREP} is utilized for extracting features of acoustic modality, including 12 Mel-frequency cepstral coefficients, pitch tracking, speech polarity, glottal closure instants, spectral envelope, etc. These acoustic features are extracted from the full audio clip of each utterance at 100Hz to form a sequence that represents variations in the tone of voice across the utterance.

We develop our model with the Pytorch framework on GTX1080Ti with CUDA 10.1 and torch 1.1.0. Our proposed model is trained with Mean Absolute Error (MAE) as loss function and with Adam \cite{Kingma2014Adam} optimizer whose learning rate is set to 0.00001. The scale factor $\lambda$ is set to 1000.

\end{document}